\pdfoutput=1

\documentclass[11pt]{article}

\usepackage{acl}

\usepackage{times}
\usepackage{latexsym}

\usepackage[T1]{fontenc}

\usepackage[utf8]{inputenc}

\usepackage{microtype}
\usepackage{listings}

\usepackage{times}
\usepackage{latexsym}

\usepackage{xcolor}
\usepackage{times}
\usepackage{ragged2e}
\usepackage{amsmath}
\usepackage{bm}
\usepackage{latexsym}
\usepackage{subfigure}
\usepackage{graphicx}
\usepackage{float}
\usepackage{longtable}
\usepackage{booktabs} 
\usepackage{multirow}
\usepackage{amssymb}
\usepackage[ruled,vlined]{algorithm2e}
\usepackage{color, soul}

\usepackage{tcolorbox}
\usepackage{cleveref}
\usepackage{todonotes}
\usepackage{amsmath}
\usepackage{booktabs}
\usepackage[export]{adjustbox}
\definecolor{codegreen}{rgb}{0,0.6,0}
\definecolor{codegray}{rgb}{0.5,0.5,0.5}
\definecolor{codepurple}{rgb}{0.58,0,0.82}
\definecolor{backcolour}{RGB}{252, 253, 246}

\lstdefinestyle{mystyle}{
    backgroundcolor=\color{backcolour},   
    commentstyle=\color{codegreen},
    keywordstyle=\color{magenta},
    numberstyle=\tiny\color{codegray},
    stringstyle=\color{codepurple},
    basicstyle=\ttfamily\footnotesize,
    breakatwhitespace=false,         
    breaklines=true,                 
    captionpos=b,                    
    keepspaces=true,                 
    numbers=left,                    
    numbersep=5pt,                  
    showspaces=false,                
    showstringspaces=false,
    showtabs=false,                  
    tabsize=2
}

\usepackage{microtype}

\title{OpenDelta: A Plug-and-play Library for Parameter-efficient \\ Adaptation of Pre-trained Models}

\author{Shengding Hu$^{1,2}$, Ning Ding$^{1,2}$, Weilin Zhao$^{1,2}$, Xingtai Lv$^1$, Zhen Zhang$^1$\\ \textbf{Zhiyuan Liu}$^{1,2,3}$\thanks{\quad corresponding author \texttt{liuzy@tsinghua.edu.cn}}, \textbf{Maosong Sun}$^{1,2}$ \\
\textsuperscript{1}Dept. of Comp. Sci. \& Tech., IAI, BNRIST, Tsinghua University, Beijing\\
\textsuperscript{2}International Innovation Center of Tsinghua University, Shanghai, China\\
\textsuperscript{3}Jiangsu Collaborative Innovation Center for Language Ability, Jiangsu Normal University\\
\texttt{hsd20@mails.tsinghua.edu.cn} \\
}

\begin{document}
\maketitle
\begin{abstract}

The scale of large pre-trained models (PTMs) poses significant challenges in adapting to downstream tasks due to the high optimization overhead and storage costs associated with full-parameter fine-tuning.
To address this, many studies explore parameter-efficient tuning methods, also framed as ``delta tuning'' in~\citet{ding2022delta}, which updates only a small subset of parameters, known as ``delta modules'', while keeping the backbone model's parameters fixed. 
However, the practicality and flexibility of delta tuning have been limited due to existing implementations that directly modify the code of the backbone PTMs and hard-code specific delta tuning methods for each PTM.
In this paper, we present OpenDelta~\footnote{GitHub Repo~\url{https://github.com/thunlp/OpenDelta}, Demo Video~\url{https://rb.gy/qjvpav}.}, an open-source library that overcomes these limitations by providing a plug-and-play implementation of various delta tuning methods. Our novel techniques eliminate the need to modify the backbone PTMs' code, making OpenDelta compatible with different, even novel PTMs. OpenDelta is designed to be simple, modular, and extensible, providing a comprehensive platform for researchers and practitioners to adapt large PTMs efficiently.


\end{abstract}

\section{Introduction}

\looseness=-1 With the rapid development of self-supervised learning methods in the realm of deep learning, especially pre-training techniques~\cite{MatthewEPeters2018DeepCW,JacobDevlin2018BERTPO,radford2018improving}, foundational pre-trained models~\cite{bommasani2021opportunities} (PTMs) have become a common cornerstone for numerous downstream tasks. And as a result, research into large-scale PTMs has flourished.

\looseness=-1 Nevertheless, the ever-expanding scale of PTMs also poses substantial obstacles in practical use. In traditional model adaptation, all the parameters of the PTMs are optimized for each downstream task, which becomes increasingly impractical as the model scales. Firstly, optimizing all the parameters incurs prohibitive computing and memory consumption; secondly, storing a fine-tuned model instance for each task or experiment significantly amplifies the storage cost.

\looseness=-1 To address these challenges, researchers have developed parameter-efficient methods for model adaptation. Such methods keep the parameters of the main model fixed and update only a small subset of parameters during adaptation. This approach, known as ``\textit{delta tuning}'', is described and surveyed in~\citet{ding2022delta}.  Different delta tuning methods have been proposed, with varying types and positions of ``\textit{delta modules}''. For example, Adapter module~\cite{houlsby2019parameter} is composed of two low-dimensional linear projection layers with an activation function, while LoRA~\cite{hu2021lora} module introduces a low-rank decomposition for the weight matrix. BitFit~\cite{zaken2021bitfit}, on the other hand, specifies the bias vector in PTMs as the delta modules. The delta module can be applied to different positions~\cite{ruckle2020adapterdrop, he2022unified, hu2022sparse} to achieve either better performance or efficiency.

\looseness=-1 Theoretically, incorporating most delta tuning methods would necessitate restructuring the backbone model, a requirement conventionally achieved through direct code manipulation.  
While this method may seem simple, it carries several disadvantages. 
Primarily, it lacks flexibility, as delta modules can theoretically be implemented in various positions, making modifications to each position in the backbone model code a cumbersome task.
Additionally, this method is not scalable, as accommodating delta tuning for newly introduced PTMs requires fresh code modifications, posing a challenge for researchers and engineers.

\looseness=-1 In this paper, we present a novel approach to implement delta tuning methods. Our approach modifies the backbone model's architecture after it is loaded into the memory. We propose four essential techniques, namely named-based addressing, dynamic tensor re-routing, runtime initialization, and a visualization system.
Using these key techniques, we build OpenDelta, an open-source toolkit for delta tuning without modifying the backbone model code. OpenDelta has several key features. 
Firstly, it is simple to use. Migrating from existing full-parameter training to delta tuning requires as few as three lines of code. For beginners or engineers, we also support automatic delta model construction. Secondly, it is modular, with delta modules implemented as independent sub-modules that can be attached to or detached from the backbone models. This feature allows different delta modules to coexist and cooperate in the same backbone model and serves multiple tasks flexibly. Thirdly, OpenDelta is highly extensible, supporting pre-trained models in a wide range of frameworks, including both official implementations from the Huggingface Library~\cite{wolf2019huggingface} and customized PTMs. It can potentially be used with newly emerged PTMs and integrated with other PTMs' frameworks for efficient training, such as the parallel training framework.

\section{Related Work}

\looseness=-1 Our work is related to delta tuning, more specifically, the implementation of delta tuning methods.


\looseness=-1 \textbf{Delta Tuning.} Delta tuning refers to the parameter-efficient method for tuning a large PTM. Different delta tuning methods~\cite{houlsby2019parameter, zaken2021bitfit, li2021prefix, hu2021lora, mahabadi2021compacter, sung2022lst} differ in both the architecture of the delta module and the positions that the delta modules are integrated into the backbone model. 
Various works have attempted to connect these disparate delta tuning approaches under a unified perspective~\cite{he2022unified, ding2022delta, hu2022sparse}. In our work, we draw inspiration from this unified viewpoint and aim to devise a framework that can support different delta tuning methods within the same pipeline. Our library includes the most popular delta tuning methods and is amenable to new methods as they emerge.

\textbf{Implementation of Delta tuning.} Previous implementation frameworks for delta tuning relied on the code modification approach. For example, AdapterHub~\cite{pfeiffer2020AdapterHub} copies a specific version of Huggingface transformers Library~\cite{wolf2019huggingface} and implement several popular delta tuning methods for a set of pre-defined PTMs. LoRA~\cite{hu2021lora} implements a limited library of LoRA linear layers. 
These methods are model-specific and involve hard-coded implementations, which restrict their usability across various PTMs. In contrast, OpenDelta represents a significant advancement as it requires no code changes to the backbone model, making it highly versatile and broadly applicable.


\section{Motivation}

\looseness=-1 
In this section, we begin by presenting the unified formulation of delta tuning. Then we underscore a set of crucial characteristics of delta tuning, focusing on the implementation aspect, which emphasizes the pressing need for a novel toolkit to aid in the research and advancement of delta tuning approaches.


\subsection{Unified Formulation of Delta Tuning} 

Although delta tuning is principally not limited to a specific type of neural networks, currently almost all the delta tuning methods are applied to PTMs~\cite{devlin2018bert, liu2019roberta, raffel2019exploring, brown2020language} with the Transformers architecture~\cite{vaswani2017attention}. 
A PTM $\mathcal{M}$ parameterized by $\Theta$ is composed of multiple sub-modules $m$, where the hidden representations $\mathbf{h}$ are passed through the sub-module to produce new hidden representation $\mathbf{h}'$, i.e., $\mathbf{h}' = m(\mathbf{h})$. 

\begin{figure*}[!htbp]
    \centering
    \includegraphics[width=0.9\linewidth]{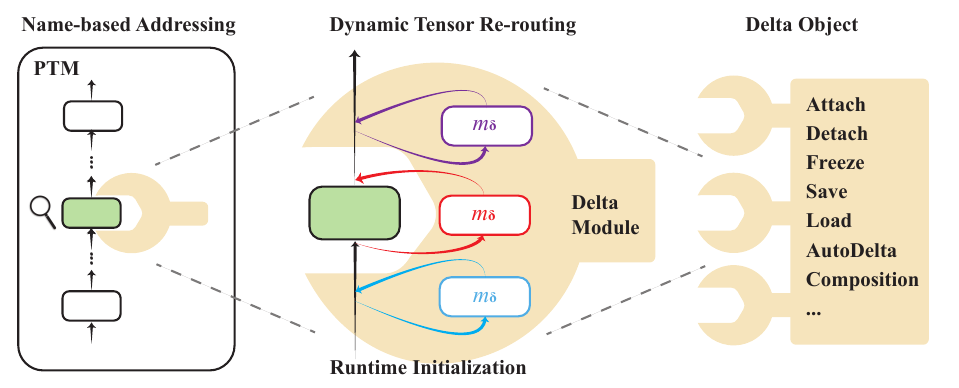}
    \vspace{-2mm}
    \caption{The overall framework of OpenDelta. The construction of delta object happens after the backbone model is loaded.}
    \label{fig:framework}
\vspace{-2mm}
\end{figure*}

The adaptation of a PTM $\mathcal{M}$ to downstream tasks is to update the original parameters $\Theta$ into $\Theta'$.  In full-parameter fine-tuning, all parameters can be updated, i.e., potentially, $|\Delta \Theta| = |\Theta|$. In contrast,  delta tuning only updates a small fraction of parameters, i.e., $|\Delta \Theta| \ll |\Theta|$. 

Despite the drastic difference in the specific form of the delta tuning methods, ~\citet{he2022unified} unify them into special forms of modifications $\Delta \mathbf{h}$ to the hidden representation $\mathbf{h}$. The $\Delta \mathbf{h}$ is generated by passing a hidden state $\mathbf{h}_{{\delta}}$ to a \textit{delta module} $m_\delta$.  Formally,
\begin{equation}
\label{equ:unifydelta}
\mathbf{h} \leftarrow \mathbf{h} + \Delta \mathbf{h} = \mathbf{h} + m_\delta(\mathbf{h}_{{\delta}}),
\end{equation}
where $\leftarrow$ denotes a replacement of the original $\mathbf{h}$, and $\mathbf{h}_{{\delta}}$ can be the same 
as or different to $\mathbf{h}$.

\subsection{Key Features for Delta Tuning} 


Several key features of delta tuning methods can be observed from Eq.(\ref{equ:unifydelta}).

\textbf{Tensor Re-routing.} The first feature of delta tuning is the ability to redirect the flow of hidden states. In a pre-trained model, the flow of hidden states forms a static graph, with the hidden states serving as nodes and sub-modules acting as transformations on the edges
As shown in Eq.(\ref{equ:unifydelta}), the introduction of the edge transformation $m_\delta$ redirects node $\mathbf{h}_{\delta}$ and injects it into another node $\mathbf{h}$, creating a new flow of hidden states that is not present in the original model architecture. The implementation of OpenDelta should achieve such tensor re-routing without hard-coding them.

\textbf{Flexibility.} 
Eq.(\ref{equ:unifydelta}) allows for the input hidden states and output hidden states to be located at any position in the backbone model $\mathcal{M}$. For example,
AdapterDrop~\cite{ruckle-etal-2021-adapterdrop} observes that only applying delta modules to the upper half of Transformer layers yields better results than the lower half. 
The flexibility of applied positions provides remarkable opportunities to explore the potential structure of delta modules~\cite{hu2022sparse}. 
However, it also presents a challenge for the implementation to be able to achieve flexibility in practice that matches the theoretical framework. 

 \textbf{Compositionality.} Different delta tuning methods can co-exist or even be combined in the same backbone model~\cite{hu2022sparse}, potentially boosting performance or supporting multitask learning~\cite{pfeiffer2020adapterfusion}.
 Thus, it is crucial to enable easy and independent implementation of each delta tuning method, while also allowing for the flexible composition of multiple modules.

 \textbf{Dynamism.} It is common for the backbone PTM to serve as a central model for multiple tasks in delta tuning. To serve a specific task, delta modules are attached to the backbone model, creating a task-specific expert. When the delta modules are detached, the backbone models revert back to their original function as general language models. This dynamic nature of delta tuning-based task adaptation should be incorporated into OpenDelta.

\section{OpenDelta}

\looseness=-1 In light of the aforementioned key features of delta tuning, we present OpenDelta. We will begin by presenting an overview of OpenDelta. Following that, we will delve into the key implementations of this framework. 

\begin{table*}
\centering
\resizebox{\textwidth}{!}{
\begin{tabular}{c|cccc}
\toprule
Method & Formulation & Default Positions & Route & Runtime Initialization  \\
\midrule
LoRA & $ m_\delta(\mathbf{h}_{\text{in}}) = \mathbf{h}_{\text{in}} \mathbf{A}\mathbf{B}$ & Query, Value & Eq.(\ref{equ:para}) & N  \\
Adapter & $ m_\delta(\mathbf{h}_{\text{out}}) = \sigma(\mathbf{h}_{\text{out}} \mathbf{W}_1)\mathbf{W}_2$ &{ATTN, FFN}  & { Eq.(\ref{equ:after}) } & {Y}  \\
Bitfit & $ m_\delta(\mathbf{h}_{\text{out}}) = \mathbf{b}$& ATTN, FFN, LayerNorm & Eq.(\ref{equ:after}) & N \\
Prefix Tuning & $ m_\delta(\mathbf{h}_{\text{out}}) = [\text{MLP}(\mathbf{p}); \mathbf{h}_{\text{out}}]$  & Key, Value & Eq.(3)  & Y \\
\bottomrule
\end{tabular}
}
\caption{Delta tuning methods and their characteristics. Default positions refer to the positions that the delta modules are attached to when no specific sub-modules are designated. $\mathbf{A}$,$\mathbf{B}$,$\mathbf{W}_1$,$\mathbf{W}_2$ are weight matrices, $\mathbf{b}$ is the bias vector.  MLP($\cdot$) is a multi-layer perception network. $[\cdot;\cdot]$ denotes the concatenation of tensors. $\sigma$ is the activation function. Runtime Initialization shows whether the implementation uses this technique in OpenDelta.}
\label{tab:defaultconfig} 
\vspace{-3mm}
\end{table*}

\subsection{Framework}

\looseness=-1 To perform delta tuning, two prerequisites are required: a pre-trained language model $\mathcal{M}$ and the ``\textit{modified modules}'', which are a user-specified list of sub-modules $m_i$ to which the delta modules should be applied. Our target is to construct a  \textit{delta object}. 
Our objective is to create a delta object, which is a collection of delta modules typically located at various positions within $\mathcal{M}$ and serves as a whole to adapt the PTM to downstream tasks. 
 We follow three steps to create a delta object.
 Firstly, we use \textit{name-based addressing} to obtain the pointers to the modified modules. Secondly, we construct a delta object comprising uninitialized delta modules.
 Thirdly, we modify the route of tensors in the modified modules into the delta modules using a \textit{dynamic tensor re-routing} technique. 
 After the updated route of the hidden state is established, we perform \textit{runtime initialization} to initialize the delta object. 
 
 \looseness=-1 After the delta object is constructed, we attach it to the backbone model. Then, we provide a simple functional interface to turn off the gradient computation in the backbone models and only compute the gradient of parameters in the delta object.
After the training is complete, we provide a simple interface for saving only the delta objects, which significantly reduces the storage requirements for the backbone model. 

\looseness=-1 The overall framework of OpenDelta is shown in Figure~\ref{fig:framework}. Next, we introduce the key implementations that support the construction of delta objects.
 

\begin{figure*}
\hspace{3mm}
\centering
\begin{minipage}{0.96\linewidth}
\begin{lstlisting}[language=Python]
model = AutoModel.from_pretrained("bert-base-cased")

+ from bigmodelvis import Visualization
+ Visualization(model).structure_graph()
+ from opendelta import LoraModel
+ delta_model = LoraModel(backbone_model=model, modified_modules=["output.dense", "query"])
+ delta_model.freeze_module(exclude=["deltas", "pooler"], set_state_dict=True)
+ Visualization(model).structure_graph()

trainer.train()
\end{lstlisting}
\end{minipage} 
\vspace{-4mm}
\caption{An example of basic usage of OpenDelta. `+' sign indicates the additional code needed to enable delta tuning. Note that the visualization can be optional if you are familiar with the backbone model.}
\vspace{-4mm}
\label{fig:basic_usage} 
\end{figure*}

 \subsection{Key Implementations}
 The above framework is achieved by four key implementations, i.e.,  name-based addressing,  dynamic tensor re-routing, runtime initialization, and visualization system.

\textbf{Name-based Addressing.} Firstly, we need to obtain a pointer to the desired sub-modules which are applied with the delta modules. In practice, we can effectively retrieve the pointer by using the name of the sub-module. Since the sub-modules are organized in a tree structure, we perform a depth-first search to find the sub-modules that match the provided name. This search results in a full path consisting of all the names from the root to the matched sub-module, accurately matching the sub-module.
However, directly writing the full path to the sub-modules can be impractical, so we design several simplifications to make addressing easier and more human-readable~\footnote{\url{https://opendelta.readthedocs.io/en/latest/notes/namebasedaddr.html}}. One such simplification involves taking advantage of the repetitiveness of transformer layers, which many delta tuning methods address by adding delta modules to the same type of sub-modules in each layer. For example, when users specify \texttt{attention}, they likely intend to apply delta modules to the attention sub-modules in all transformer layers. To address this need, we provide a tail-matching mechanism that automatically matches the sub-modules based on their names. For more complex configurations of positions, we allow matching based on regular expressions and web-based selection using our custom-designed web interface.



\textbf{Dynamic Tensor Re-routing.} A fundamental distinction that sets OpenDelta apart from other implementations is its ability to add delta modules without requiring any modifications to the code of the backbone modules. This feature necessitates a dynamic rerouting of tensors through the delta modules and back into the backbone model.
To achieve this rerouting, we wrap the original forward function of a sub-module with a wrapper function and replace the original forward function with the wrapper function. To ensure seamless replacement, we utilize a decorator to inherit the original function's attributes, including the I/O, doc string, etc. Within the wrapped function, we implement three distinct routes of the hidden states, taking into account the order of the original sub-module and the delta module.
The first route utilizes the input hidden state $\mathbf{h}_\text{in}$ of $m_i$ as both the modification target and the input to the delta module. We pass it through the delta module to get the output $m_\delta(\mathbf{h}_\text{in})$, and merge it to $\mathbf{h}_\text{in}$. Formally, 
\begin{equation}
\label{equ:in}
\mathbf{h}_\text{in}\leftarrow\mathbf{h}_\text{in} + m_\delta(\mathbf{h}_\text{in}).
\end{equation}
The second route employs the output hidden state $\mathbf{h}_\text{out}$ of $m_i$  as the modification target:
\begin{equation}
\label{equ:after}
\mathbf{h}_\text{out}\leftarrow\mathbf{h}_\text{out} + m_\delta(\mathbf{h}_\text{out}).
\end{equation}
The third route leverages the input hidden state $\mathbf{h}_\text{in}$ as the input to the delta module, and sets the output hidden state $\mathbf{h}_\text{out}$ as the modification target: 
\begin{equation}
\label{equ:para}
\mathbf{h}_{\text{out}}\leftarrow\mathbf{h}_\text{out} + m_\delta(\mathbf{h}_\text{in}).
\end{equation}

\looseness=-1 While these three routes do not necessarily encompass all possible relationships between the delta module and the backbone model, they are sufficient to support most popular delta tuning methods (as illustrated in Table~\ref{tab:defaultconfig}). However, we remain open to the possibility of incorporating additional routes as needed.

\looseness=-1 \textbf{Runtime Initialization.} To ensure that weight matrices in the delta module match the hidden states in terms of shape and dimension, we must account for hidden states whose shapes are not specified in the model configuration. In traditional implementations, this requires manually examining the code of the backbone model. However, OpenDelta automates this process by passing a pseudo input through the backbone model, allowing the shapes of the hidden states to be automatically determined as they propagate from the input to the output.

\begin{figure}
    \centering
    \includegraphics[width=\linewidth]{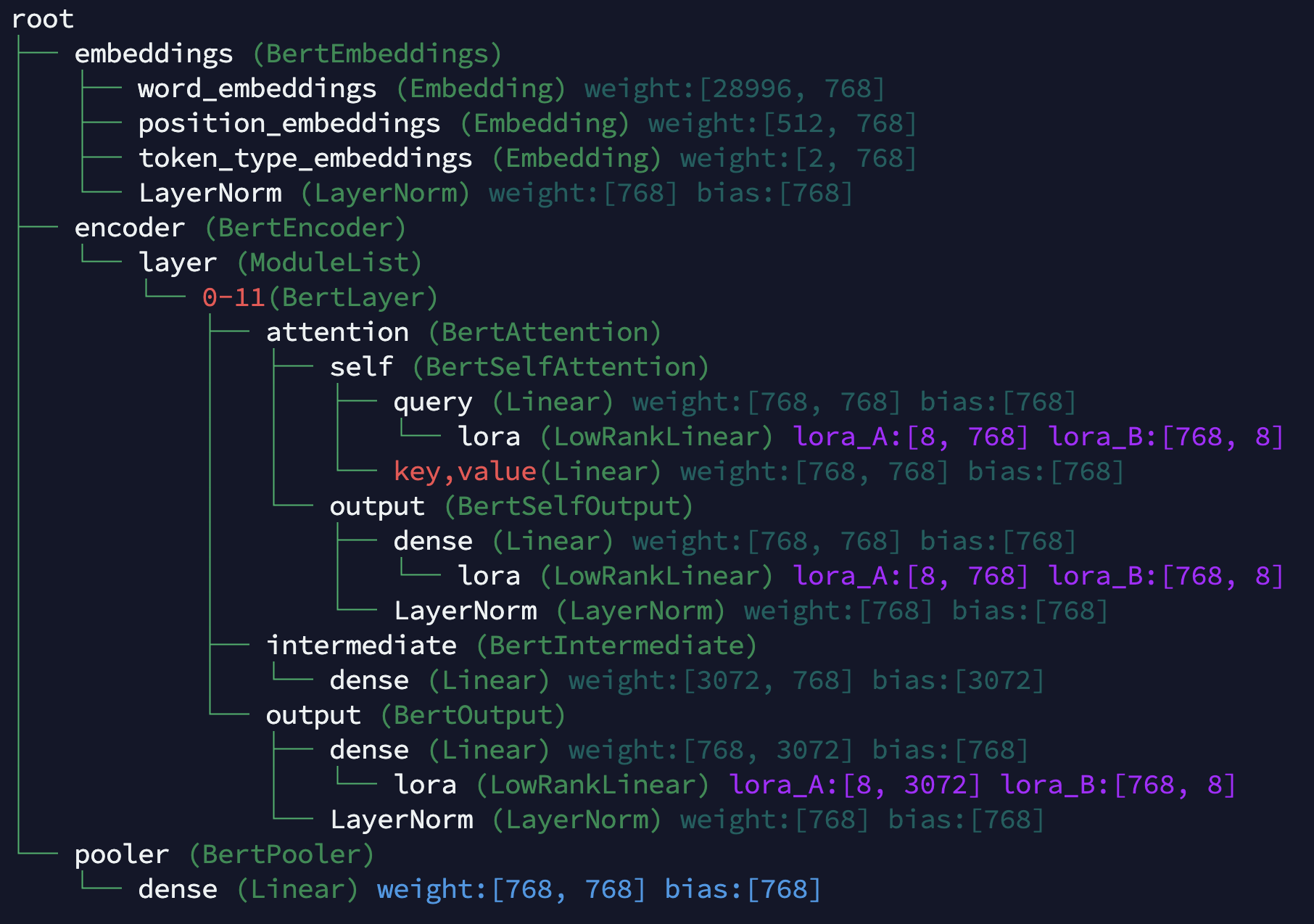}
    \caption{The visualization of the backbone model's status after the LoRA modules are attached.}
    \label{fig:after_lora}
    \vspace{-3mm}
\end{figure}

\looseness=-1 \textbf{Visualization System.} As delta tuning provides flexibility and dynamism, it is essential to ensure the correct construction of delta objects by verifying that delta modules are added as specified. However, direct printing of large pre-trained models results in massive outputs.
To address this, we provide a visualization system that leverages repetition in transformer architecture. Specifically, we collapse the repetitive layers and neatly print the parameters' information. With the addition of delta modules to the backbone model, users can easily observe the changes made in the model through visualization. An example of visualization can be seen in Figure~\ref{fig:after_lora}. As the visualization system is useful beyond delta tuning, it has been separated into an independent package named ``\texttt{bigmodelvis}''~\footnote{\url{https://pypi.org/project/bigmodelvis/}}.

a

\section{Usage}

In this section, we provide the use cases of OpenDelta which demonstrate the three characteristics of OpenDelta, i.e., simplicity, modularity, and extensibility. 

\begin{figure*}[hbt!]
\vspace{2mm}
\centering
\begin{minipage}{0.94\linewidth}
\begin{lstlisting}[language=Python]
def multi_task(delta_model, input_text):
    global model # We use the same backbone model across tasks. 
    delta_model.attach()
    print(tokenizer.decode(model.generate(input_ids=tokenize(input_text))))
    delta_model.detach()
multi_task("What the commmon career of Newton ad einstein?", spelling_delta)
# >>> "What was the common career of Newton and Einstein?"
multi_task("What was the common career of Newton and Einstein?", topic_delta)
# >>> "The question's topic is science."
multi_task("What was the common career of Newton and Einstein?", question_delta)
# >>> "Physicists."


\end{lstlisting}
\end{minipage} 
\vspace{-0.4cm}
\caption{Multitask learning via OpenDelta. Due to space limitations, we retain only the core code. For detailed code, please refer to the OpenDelta documentation. Strings after ``> > >'' demonstrate the output of the model. }
\vspace{-0.5cm}
\label{fig:code_multi_task} 
\end{figure*}

\subsection{Simplicity}

\textbf{Migrating from Fine-tuning.} To facilitate the migration from existing full-parameter fine-tuning to delta tuning, only a few lines of code modifications are required, as exemplified in Figure~\ref{fig:basic_usage}. Initially, in the traditional full-parameter fine-tuning, the PTM is loaded from external libraries, such as Huggingface Transformers (Line 1), and train the model (Line 10). To introduce delta tuning, line 3-8 are added and executed. To begin with, an optional step is to visualize the backbone model to identify the target ``\texttt{modified\_modules}''. Then, a delta object, such as LoRA, is created and attached to the backbone model. Subsequently, the model parameters, excluding the delta modules and the randomly initialized classification head, are frozen. The ``\texttt{set\_state\_dict=True}'' parameter is employed to remove the non-trainable parameters from the model checkpoint. Lastly, the sub-modules of the backbone are visualized to verify the successful creation and attachment of the delta modules. An example of the visualization results is depicted in Figure~\ref{fig:after_lora}.


\textbf{AutoDelta Mechanism.} The implementation of OpenDelta supports highly intricate designs of delta modules, catering to diverse experimental requirements. Nonetheless, it is desirable to provide a default configuration of delta modules for practitioners who may not be well-versed in the mechanism of delta tuning. 
However, the naming conventions of sub-modules differ significantly among various backbone models, despite their shared transformer architecture. 
To tackle this issue, we establish a common name convention and employ a mapping technique to map the model-specific name convention to the common one~\footnote{\url{https://opendelta.readthedocs.io/en/latest/notes/unifyname.html}}. This enables the AutoDelta mechanism to be supported seamlessly.
Figure~\ref{fig:autodelta} exemplifies that, once the type of the delta tuning method is specified, the delta modules will be attached to the backbone model in default positions and with appropriate hyper-parameters. We have listed the default configurations of each delta tuning method in Table~\ref{tab:defaultconfig}. Furthermore, the AutoDelta mechanism facilitates the loading of fine-tuned checkpoints of delta modules, without explicit knowledge of the type and hyper-parameters of the delta modules.

\begin{figure}[hbt!]
\hspace{2mm}
\vspace{-3mm}
\centering
\begin{minipage}{0.94\linewidth}
\begin{lstlisting}[language=Python]
from opendelta import AutoDeltaModel, AutoDeltaConfig
# construct a new delta using the default configuration.
delta_config = AutoDeltaConfig.from_dict({"delta_type":"lora"})
delta_model = AutoDeltaModel.from_config(delta_config, backbone_model)
# load the delta checkpoint.
delta = AutoDeltaModel.from_finetuned("save_dir", backbone_model)
\end{lstlisting}
\end{minipage}
\vspace{-2mm}
\caption{An example of using AutoDelta mechanism.}
\vspace{-5mm}
\label{fig:autodelta}
\end{figure}


\subsection{Modularity}
The second notable attribute of OpenDelta is modularity. It affords the capacity to independently attach and detach each delta object from the backbone model, thereby providing the possibility of multi-task serving with a single backbone model. Specifically, suppose data pertaining to various tasks are presented sequentially, wherein each data triggers the attachment of a corresponding delta object to the backbone model for processing, and once completed, the delta object is detached. A case that illustrates this functionality is illustrated in Figure~\ref{fig:code_multi_task}, where three tasks are process sequentially using a single backbone model. 


\subsection{Extensibility}
Delta tuning is one of the important techniques that enables the use of large PTMs, and as such, we make efforts to ensure its compatibility with other techniques such as model acceleration and multi-GPU training. Specifically, we currently provide support for the BMTrain framework~\footnote{\url{https://github.com/OpenBMB/BMTrain}} with ZeRO-3 optimization enabled~\cite{rajbhandari2020zero}. It is also worth noting that we plan to expand our support for additional model-acceleration frameworks in the future.

\section{Conclusion}
In summary, OpenDelta is a plug-and-play library for delta tuning, offering an intuitive and modular solution to adapt large PTMs using delta tuning without the need for code modifications. The library's user-friendliness, flexibility, and extensibility make it accessible and useful for both researchers and engineers. In the future, we plan to continuously update the library with new delta tuning methods and ensure its compatibility with the latest versions of other major PTMs libraries.

\newpage

\section{Acknowledgements}
This work is supported by the National Key R\&D Program of China (No.2022ZD0116312), National Natural Science Foundation of China (No. 62236004), Major Project of the National Social Science Foundation of China (No. 22ZD298).

\section*{Limitations}
Although we believe that OpenDelta is simple, easy to use, flexible, and extensible since it does not require code modification, it is still limited by many implementation details. For example, some delta tuning methods, such as Prefix Tuning, are limited by theory and can only be used in Attention layers, making them unable to be arbitrarily specified. This is also why we did not use it as an example in this paper. On the other hand, some base models differ significantly from mainstream implementations, making it difficult to use the AutoDelta mechanism. Therefore, we maintain a list of tested models that can use AutoDelta, while other models may still use OpenDelta in a customized manner. Thirdly, while theoretically compatible with acceleration frameworks other than BMTrain, such as Deepspeed, there are some implementation details that currently limit the compatibility of some functions. We will do our best to communicate with the maintainer of those packages to increase compatibility.

\section*{Ethical Consideration}
In the writing process of this paper, ChatGPT~\cite{chatgpt} was utilized for revision and refinement. However, the authors can guarantee that each sentence in this paper has been thoroughly reviewed and checked to accurately convey the authors' intended meaning.

\bibliography{anthology,custom}
\bibliographystyle{acl_natbib}

\clearpage





\end{document}